\documentclass[letterpaper, 10 pt, journal,twoside]{./iEEEtran}
\usepackage{amsmath,amssymb,amsfonts}
\usepackage{balance}
\usepackage{bm}
\usepackage{hyperref}
\usepackage{color}
\usepackage[table,xcdraw]{xcolor}
\usepackage{tabularx}
\usepackage[caption=false,font=normalsize,labelfont=sf,textfont=sf]{subfig}
\usepackage{booktabs}
\usepackage{graphicx}
\usepackage{multirow}
\usepackage{tikz}
\usepackage{pgfplots}
\usepackage{siunitx}
\usetikzlibrary{bending, external}

\graphicspath{{./Figures/}}
\DeclareGraphicsExtensions{.pdf,.png,.jpg,.eps,.svg}

\IEEEoverridecommandlockouts
\pgfplotsset{compat=1.16} 

\title{Predictor-Based Time Delay Control of A Hex-Jet Unmanned Aerial Vehicle}

\author{Junning Liang$^{1}$, Haowen Zheng$^{1}$, Yuying Zhang$^{1}$, Yongzhuo Gao$^{2}$, Wei Dong$^{2}$, Ximin Lyu$^{1}$
    \thanks{Manuscript received: Oct, 8th, 2024; Revised Dec, 5th, 2024; Accepted Jan, 30th, 2025. 
    This paper was recommended for publication by Editor Giuseppe Loianno upon evaluation of the Associate Editor and Reviewers' comments.     
    This work is supported by the Shenzhen Excellent Scientific and Technological Innovation Talent Training Program (Grant No. RCBS20221O08093104017)  and the National Natural Science Foundation of China (Grant No.62303495). 
    (\textit{Junning Liang and Haowen Zheng contributed equally to this work.}) (\textit{Corresponding author:} Ximin Lyu.)}
    \thanks{$^1$ The authors are with the School of Intelligent Systems Engineering, Sun Yat-sen University, Guangzhou, China 
    
    (e-mail: {liangjn33@mail2.sysu.edu.cn; zhenghw23@mail2.sysu.edu.cn; zhangyy393@mail2.sysu.edu.cn; lvxm6@mail.sysu.edu.cn}).}
    \thanks{$^2$ The authors are with the State Key Laboratory of Robotics and Systems, Harbin Institute of Technology, Harbin, China.}
    \thanks{Digital Object Identifier (DOI): see top of this page.}
}

\begin{document}
\maketitle

\begin{figure*}[t]
        \vspace{0pt}
	\centering
	\includegraphics[width=1\textwidth]{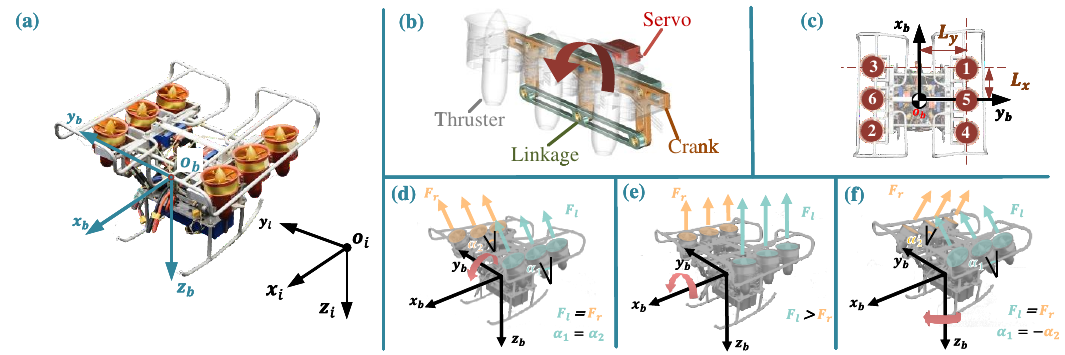}
	\caption{(a).~The scaled Hex-Jet and its coordinate system: blue arrows indicate the body frame and black arrows stand for the inertial frame. (b).~3D model of the Hex-Jet`s parallelogram uniaxial thrust vectoring mechanism. 
    (c).~Schematic view of the scaled Hex-Jet. The numbers in the circle indicate the index of the turbojet. $L_x$ and $L_y$ stands for the distance between the turbojet and the center of gravity of the vehicle.
    (d).~Hex-Jet differential thrust for pitch rotation. The blue and red arrows represent the magnitude and direction of thrust on the left and right sides, respectively. (e).~Hex-Jet thrust vectoring for roll rotation. (f).~Hex-Jet thrust vectoring for yaw rotation.
	\label{fig:sys_overview}}
	\vspace{0pt}
\end{figure*}

\begin{abstract}

Turbojet-powered VTOL UAVs have garnered increased attention in heavy-load transport and emergency services, due to their superior power density and thrust-to-weight ratio compared to existing electronic propulsion systems. 
The main challenge with jet-powered UAVs lies in the complexity of thrust vectoring mechanical systems, which aim to mitigate the slow dynamics of the turbojet. 
In this letter, we introduce a novel turbojet-powered UAV platform named Hex-Jet.
Our concept integrates thrust vectoring and differential thrust for comprehensive attitude control. 
This approach notably simplifies the thrust vectoring mechanism. 
We utilize a predictor-based time delay control method based on the frequency domain model in our Hex-Jet controller design to mitigate the delay in roll attitude control caused by turbojet dynamics.
Our comparative studies provide valuable insights for the UAV community, and flight tests on the scaled prototype demonstrate the successful implementation and verification of the proposed predictor-based time delay control technique.

\end{abstract}


\section{INTRODUCTION}	

\IEEEPARstart{I}{n} recent years, there has been a remarkable surge of interest in aerial transportation \cite{thiels2015use}.
Multirotors have gained prominence for such missions due to their mechanical simplicity, robustness, and vertical take-off and landing capability.
However, as payload requirements increase, multirotors encounter effective size-related challenges, larger propellers are necessary to accommodate heavier payloads~\cite{leishman2006principles}.
This leads to bulky multirotor designs, rendering them unsuitable for confined environments. Moreover, large propeller blades pose substantial safety risks to individuals and environments in urban areas.

To address the above issues, researchers have started exploring the potential of turbojet engines.
G L'Erario \textit{et al}.~\cite{l2020modeling} conducted the first systematic study on the application of model turbojet engines in aerial humanoid robotics.
Compared to rotor-based propulsion systems, turbojet engines offer a higher thrust-to-weight ratio with a smaller effective size, avoiding direct exposure of rotor blades\cite{alulema2021propulsion}, and providing enhanced safety for human interaction.
In the unmanned aerial vehicle community, researchers have explored the utilization of turbojet-engine-powered vertical takeoff and landing aircraft.

Compared to the multirotors, UAVs based on turbojet engines typically require additional thrust vectoring mechanisms for attitude control, due to the relatively slow response of turbojet engines and the smaller counter-torque effects. 
By varying the thrust component of each actuator in different directions, thrust vectoring actuators can provide the desired control force and torque.
These thrust vectoring mechanisms include the incorporation of thrust vectoring nozzles~\cite{tang2022modeling,yin2023novel} or turbojet thrust vectoring mechanisms~\cite{turkmen2020design} to provide rapid control torque.

There are two types of primary mechanisms used for thrust vectoring control in jet-powered UAVs: uniaxial thrust vectoring~\cite{tang2022modeling,yin2023novel} and bi-axial thrust vectoring~\cite{turkmen2020design}.
In a uniaxial thrust vectoring vehicle, the actuator is rotated about a single axis. 
However, this mechanism also introduces power inefficiencies due to the partial cancellation of non-parallel thrusts.
In contrast, a bi-axial thrust vectoring vehicle avoids the inefficiencies caused by opposite thrusts by ensuring that the thrust directions remain parallel~\cite{zheng2020tiltdrone}.
On the other hand, these designs come with a weight penalty and increased mechanical complexity.

Differential thrust, a straightforward method for tilt attitude control, is widely utilized in multirotor UAVs.
This approach eliminates the need for intricate thrust vectoring mechanisms but demands swift responsiveness from the actuators. 
However, turbojet-powered UAVs can't directly use this driven method without appropriate controllers due to the large time delay of the turbojet.

As an effective way to handle the time delay system, the time delay control methods have been widely applied in the industrial community~\cite{deng2022predictor,da2020controlling}.
In the field of unmanned aerial vehicles, R. Lozano \textit{et al}.~\cite{lozano2004robust} discussed the measurement delay problem in the UAV's control system for the first time. 
They used the d-step ahead state predictor to solve the input delay problem caused by the measurement and computation of the controller input. 
This approach was validated through experiments on the yaw control of a mini-quadrotor. 
Expanding on this concept, R. Sanz \textit{et al}.~\cite{sanz2016predictor} extended the idea to encompass both measurement and actuator delays in the design of uncertainty and disturbance estimators (UDEs). 
Additionally, Y. Yu \textit{et al}.~\cite{yu2015high} and Z. Qin \textit{et al}.~\cite{qin2023design} used the Smith predictor to address the actuator delay in control allocation design. 
However, related work~\cite{sanz2016predictor,yu2015high,qin2023design} primarily focuses on addressing relatively short actuator delays (25-60ms) or large communication delays~\cite{lee2022robust} in small multirotor UAVs. 
There is a lack of research on the longer actuator delays in unmanned aerial vehicle fields.

In this letter, we propose a new turbojet-powered aircraft design called Hex-Jet. 
We utilize the predictor-based time delay control method to overcome the challenge posed by the slow response of turbojet engines. This approach effectively compensates for the delayed response of the turbojet engines, ensuring satisfactory control performance of the system.

To be specific, we summarize our key contributions as follows:
\begin{enumerate}
\item \textbf{Hex-Jet}: We introduce a novel type of turbojet-powered aerial vehicle platform called Hex-Jet. Combining uniaxial thrust vectoring and differential thrust, our platform features a simpler mechanical structure, and reduced weight compared to existing jet-powered UAV configurations.
    
\item \textbf{Comparative study of predictor-based time delay control}: 
To select a predictor-based controller that meets the performance requirements of Hex-Jet flight control, we conduct a series of comparative studies of two different predictor-based time delay control approaches on a quadrotor platform. 
In our experiment results, we observed that the Smith predictor-based controller exhibits superior disturbance rejection performance and robustness to modeling errors compared to the state predictor-based controller. 
The comparative experimental results also provide valuable insights for other UAV time delay control applications considering the adoption of these two methods.

\item \textbf{Successful implementation and verification on the scaled Hex-Jet platform}: We successfully implemented and verified the Smith predictor-based controller on our scaled Hex-Jet, considering a specific time delay. 
Compared to the baseline controller, the Smith predictor-based controller can stabilize our prototype vehicle successfully with desirable tracking performance.
To the best of our knowledge, this is the first introduction of time delay control methods to turbojet-powered UAVs. 
\end{enumerate}

\section{MODELING AND BASELINE CONTROLLER}
\label{sec:modeling}

\subsection{Coordinate Frames and System Design}
\label{subsec:sysconfig}

We introduce two coordinate frames of our Hex-Jet UAV for further discussion and they are shown in Fig.~\ref{fig:sys_overview}(a): the body frame $\bm{x}_b$, $\bm{y}_b$, $\bm{z}_b$, and the inertial frame $\bm{x}_i$, $\bm{y}_i$, $\bm{z}_i$. 
The body frame's origin $\bm{o}_b$ is set to coincide with the vehicle's center of gravity. 
Rotation along body axis $\bm{x}_b$, $\bm{y}_b$, and $\bm{z}_b$ are respectively called roll, pitch, and yaw (denoted by $\phi$, $\theta$, and $\psi$).

The system configuration of the Hex-Jet proposed in this letter is shown in Fig.~\ref{fig:sys_overview}(a). 
All actuators are mounted on a solid aluminum frame.
Essential avionics, including the Engine Control Unit (ECU) and the Flight Control Unit (FCU), are mounted within the frame. 
The Hex-Jet consists of two parallelogram uniaxial thrust vectoring mechanisms, two servos, and six model turbojet engines as the dominant propulsion system. 
The key specifications for our scaled Hex-Jet are illustrated in Table.~\ref{tab:scaled_hexjet_spec}.

The uniaxial thrust vectoring mechanism is shown in Fig.~\ref{fig:sys_overview}(b).
The mechanism consists primarily of three cranks, a linkage, and a servo. 
Three turbojet engines are mounted in parallel on each parallelogram uniaxial thrust vectoring mechanism, which is driven by a single servo. 
The servo drives only the central crank, the lower linkage and other driven cranks form a parallelogram mechanism. 
Therefore, the three turbojets on each side rotate at an identical angle.
By adjusting the angle of the servos on both sides, we can maneuver the UAV in the pitch and yaw axes. 
As shown in Fig.~\ref{fig:sys_overview}(e) and (f).
The roll axis of the vehicle can be driven by differentiating the total thrust generated by the turbojet engines on either side of the thrust vectoring mechanisms as shown in Fig.~\ref{fig:sys_overview}(d).

\begin{table}[t]
    \small
    \centering
    \caption{Scaled Hex-Jet Key Specifications}
    \label{tab:scaled_hexjet_spec}
    \resizebox{\columnwidth}{!}{
    \begin{tabular}{ll}
        \toprule
        Flight Control Unit & STM32F4 \\
        Dimensions (length {$\times$}width{$\times$}height) & 400{$\times$}400{$\times$}260 mm\\
        Maximum Thrust & 12.0 kg\\
        Takeoff Weight & 5.525 kg\\
        \bottomrule
    \end{tabular}
    }
\end{table}

We denote different numbers to distinguish each turbojet: $i\ {\in}\ \{1,\ 2,\ 3,\ 4,\ 5,\ 6\}$, as shown in Fig.~\ref{fig:sys_overview}(c).
Turbojet 1, 4, 5 and 2, 3, and 6 are grouped together respectively, their tilt angle can only be changed simultaneously. 
We represent the tilt angle of turbojets 2, 3, and 6 as ${\alpha}_1$, and the tilt angle of turbojets 1, 4, and 5 as ${\alpha}_2$. 
${\alpha}_1$ and ${\alpha}_2$ also represent the rotation angles of the left servo and the right servo, respectively. 
We use the right-hand rule to determine the positive direction of each servo's rotation around the y-axis in the body frame. 
$\bm{L}_{i}$ denotes the distance from the center of gravity to $i$-th engine in the body frame, see the red arrows in Fig.~\ref{fig:sys_overview}(c), which is given by
\begin{equation}
\small{
    \begin{matrix}
        \bm{L_{1}} = \begin{bmatrix} \ \ L_{x} & \ \ L_{y} & L_{z} \end{bmatrix}^{T}, & 
        \bm{L_{2}} = \begin{bmatrix} -L_{x} & -L_{y} & L_{z} \end{bmatrix}^{T},\\
        \bm{L_{3}} = \begin{bmatrix} \ \ L_{x} & -L_{y} & L_{z} \end{bmatrix}^{T},&
        \bm{L_{4}} = \begin{bmatrix} -L_{x} & \ \ L_{y} & L_{z} \end{bmatrix}^{T},\\
        \bm{L_{5}} = \begin{bmatrix} \ \ 0 & \ \ \ \ L_{y} & L_{z} \end{bmatrix}^{T},&
        \bm{L_{6}} = \begin{bmatrix} \ \ 0 & \ \ -L_{y} & L_{z} \end{bmatrix}^{T}.
    \end{matrix}
}
\end{equation}

\subsection{Modeling}
\label{subsec:modeling}

We formulate the equation of motion for the Hex-Jet UAV as follows
\begin{equation}
    \begin{aligned}
        m\dot{\bm{v}}_{i} &= \bm{f}_{ig} + \mathbf{R}\bm{F}_{b}\\
        \mathbf{J}\dot{\bm{\omega}}_{b} &= \bm{M}_{b} - \bm{\omega}_{b}{\times}\mathbf{J}\bm{\omega}_{b},
    \end{aligned}
\end{equation}
where $m$ denotes the UAV's total mass, $\bm{v}_i$ is the vehicle velocity in the inertial frame, $\bm{f}_{ig}$ is gravitational force and $\mathbf{R}$ is rotation matrix from the body frame to inertial frame, $\bm{F}_b$ is the force generated by the turbojet in the body frame. 
$\mathbf{J}$ is the inertia matrix, $\bm{\omega}_{b}$ is the angular rate in the body frame, $\bm{M}_{b}$ is the torque generated by the turbojet in the body frame.

The total force produced by the $i$-th turbojet in the body frame rotated by two servos can be assembled by
\begin{equation}
\bm{F}_{b} = \sum^{6}_{i=1}\bm{F}^{i}_{b} = 
    \begin{bmatrix}
	3F^{i}_{b}(\text{sin}(\alpha_{1}) + \text{sin}(\alpha_{2}))  \\
	0                                      \\
	3F^{i}_{b}(\text{cos}(\alpha_{1}) + \text{cos}(\alpha_{2}))
    \end{bmatrix}.    
\end{equation}

The torque produced by the turbojets can be divided into two parts: the pitch and yaw torque come from the misalignment between the thrust vector and center of gravity by rotating the thrust vectoring mechanism. 
The roll torque comes from the differential thrust between the left and right group of turbojets.

During the pitch and yaw maneuvering, the rotation relation of the servo holds ${\alpha}_1 ={\alpha}_2$ for pitch, and ${\alpha}_1 = -{\alpha}_2$ for yaw. 
Then, we can obtain the y and z-axis torque by
\begin{equation}\label{eq:yz_torque}
\small{
\bm{M}^{yz}_{b} = \begin{bmatrix}
    0 \\
    3*(F^{i}_{b}*L_{z}*\text{sin}({\alpha}_1) + F^{i}_{b}*L_{z}*\text{sin}({\alpha}_2)) \\
    3*(F^{i}_{b}*L_{y}*\text{sin}({\alpha}_2) - F^{i}_{b}*L_{y}*\text{sin}({\alpha}_1)) 
\end{bmatrix}.
}
\end{equation}
The roll axis control moment is generated by differential thrust. 
Since the servo has only two cases of ${\alpha}_1 ={\alpha}_2$ and ${\alpha}_1 = -{\alpha}_2$, the vertical component in the body frame remains constant at all times.
Denote the thrust of each turbojet on the left and right to $\bm{F}_L$ and $\bm{F}_R$, we can derive the x-axis torque:
\begin{equation}\label{eq:x_torque}
\bm{M}^{x}_{b} = \begin{bmatrix}
    3(F_R*L_{y} - F_L*L_{y}) \\
    0 \\
    0 
\end{bmatrix}.    
\end{equation}
Combining~\eqref{eq:yz_torque} and~\eqref{eq:x_torque}, We can obtain full attitude control torque
\begin{equation}
\bm{M}_{b} = \bm{M}^{x}_{b} + \bm{M}^{yz}_{b}.    
\end{equation}
\subsection{Baseline Controller Structure}
\label{subsec:controller}

\begin{figure}[h]
	\vspace{-10pt}
	\centering
	\includegraphics[width=1.0\columnwidth]{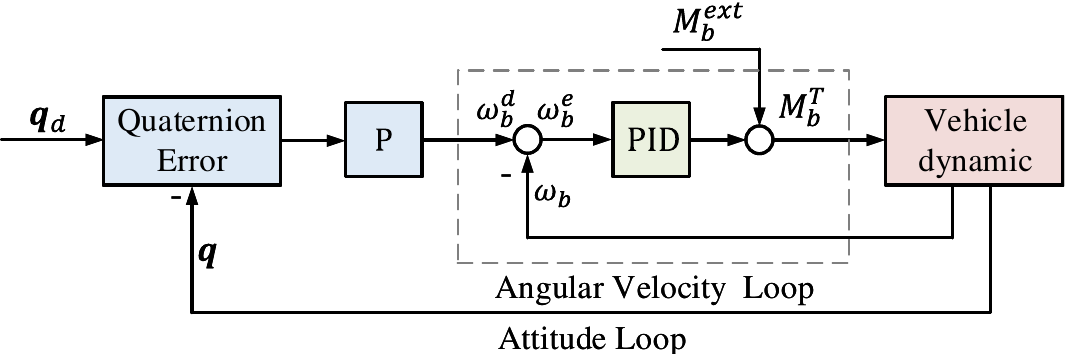}
	\caption{The attitude controller diagram.
		\label{fig:attitude_controller_diagram}}
\end{figure}

In this letter, we focus on the attitude control loop of the vehicle. 
We use a cascade control structure, as shown in Fig.~\ref{fig:attitude_controller_diagram}. 
The attitude controller is a proportional controller based on the quaternion error, and the rate controller is a Proportional-Integral-Derivative controller.

\begin{figure}[t]
	\centering
        \includegraphics[width=1.0\columnwidth]
    {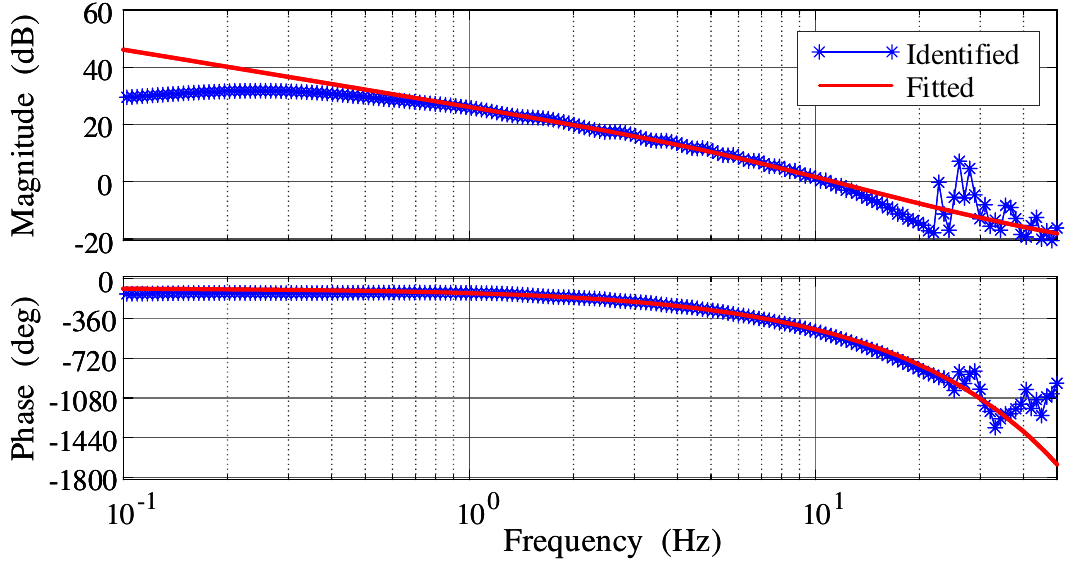}
	\caption{System identification results, line with asterisk marker present frequency response from sweep data, solid line represent the fitted result from frequency response.
		\label{fig:sys_identification}}
	\vspace{-10pt}
\end{figure}

 \subsection{System Identification and Scalability}
\label{subsec:system_plant}

We identify the turbojet thrust dynamic model in the frequency domain by frequency sweep methods\cite{pintelon2012system}. 
The transfer function of the turbojet is shown in~\eqref{eq:turbo_model}, which is a first-order transfer function with a time delay.

\begin{equation}\label{eq:turbo_model}
   G_{turbo}(z) =\frac{0.01866}{z - 0.9879}{z}^{-25}.   
\end{equation}

Considering a sampling time of 0.004s, the transfer function $G_{turbo}(z)$ represents a time constant of 0.33 seconds and a delay of 0.1 seconds (25 samples) in the turbojet.
Compared to the transfer function model of the electric ducted fans (EDF) used in our scaled Hex-Jet, as shown in~\eqref{eq:edf_plant}, the time constant of the EDF is smaller, which is 0.029s.

\begin{equation}\label{eq:edf_plant}
    G_{EDF}(z) = \frac{0.2554}{z-0.8748}.
\end{equation}

Under identical delay conditions, the scaled Hex-Jet exhibits higher sensitivity than its full-scale counterpart due to the higher control bandwidth required for smaller aircraft~\cite{pounds2010modelling}.
To align our scaled Hex-Jet more closely with the full-scale model, we integrated the transfer function $G_{turbo}(z)/G_{EDF}(z)$ into our scaled model for frequency domain identification. This incorporation resulted in a refined transfer function model that better represents the dynamics of the turbojet-powered vehicle.

Corrected by the turbojet dynamics~\eqref{eq:turbo_model}, the refined roll rate response results in the frequency domain are shown in Fig.~\ref{fig:sys_identification}.
We use a second-order plus input delay transfer function to fit the roll channel frequency response result of our scaled Hex-Jet dynamics yielding the following transfer function:
\begin{equation}
    G(z) = \frac{0.0689(z+1.4538)}{(z-0.9999)(z-0.9024)}{z}^{-25}.
\end{equation}

\section{PREDICTOR BASED CONTROLLER DESIGN}
\label{sec:predictor_design}

In this section, we introduce two types of predictor-based time delay controllers. 
Based on our Hex-Jet rate controller structure, we design the predictor-based rate controller using the identified frequency domain model.

\subsection{Smith Predictor}
\label{subsec:smith_predictor}

\begin{figure}[t]
        \vspace{2pt}
	\centering
	\includegraphics[width=1\columnwidth]{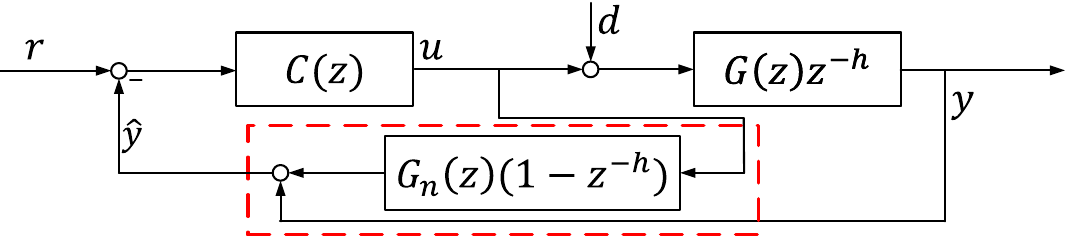}
	\caption{Smith Predictor structure.
		\label{fig:smith_predictor}}
	\vspace{-0pt}
\end{figure}

Smith predictor\cite{smith1957closer} is the first reported predictor-based controller to regulate a delay system by introducing an inner system acting as a predictor.
By constructing an inner loop using the Smith predictor, as shown in Fig.~\ref{fig:smith_predictor}, we can make an equivalent closed loop with a delay-free system $G(z)$, and stabilize it with the baseline controller $C(z)$.

The Smith predictor in the inner loop can be expressed as
\begin{equation}
    P(z) = G_{n}(z) - G_{n}(z)z^{-h},
\end{equation}
such that the predicted output $\hat{y}(z)$ can be expressed by
\begin{equation}
    \hat{y}(z
    ) = P(z)u(z) + y(z).
\end{equation}

Replace $y$ by $\hat{y}$, we can formulate the equivalent inner loop system
\begin{equation}\label{eq:eqv_cl_loop}
    \frac{Y(z)}{R(z)} = \frac{C(z)N(z)}{D(z)+C(z)N(z)}z^{-h},
\end{equation}
where $N(z)$ is the numerator and $D(z)$ is the dominator of the delay-free system $G(z)$.
We can see that the delay term of the system can not influence the closed-loop stability of \eqref{eq:eqv_cl_loop}.

\subsection{D-step Ahead State Predictor}
\label{subsec:state_predictor}
The d-step ahead state predictor is used to predict the state of the system, while the Smith predictor acts as an output predictor.
The concept of state prediction was initially introduced by Manitius \textit{et al}.~\cite{manitius1979finite}, with subsequent digital implementation discussed by Goodwin \textit{et al}.\cite{goodwin2014adaptive}. 
Lozano~\textit{et al}.~\cite{lozano2004robust} proposed a d-step ahead state predictor-based control scheme for linearized quadrotor yaw control systems with uncertainties. 
In this letter, we design the angular rate state predictor with the state observer using the frequency domain model.

We define $\bm{x}^p_{k+d}$ as the prediction of the state $\bm{x}_{k+d}$ at time sample $t_k$, the discrete form of the d-step ahead state predictor can be expressed as
\begin{equation}\label{eq:state_predictor}
\bm{x}^p_{k+d} = \mathbf{A}^{d}\bm{x}_{k} + \mathbf{A}^{d-1}\mathbf{B}\bm{u}_{k-d} + \cdots + \mathbf{B}\bm{u}_{k-1}.   
\end{equation}
The predicted state $\bm{x}^p_{k+d}$ can be computed using system state available at time $t_k$ and previous input $\bm{u}_{k-1}, \bm{u}_{k-2}, ..., \bm{u}_{k-d}$.
By implementing a suitable controller, we can stabilize the plant $[\mathbf{A},\mathbf{B}]$ with robustness against minor uncertainties in the system.

\begin{figure}[t]
	\vspace{-0pt}
	\centering
	\includegraphics[width=1\columnwidth]{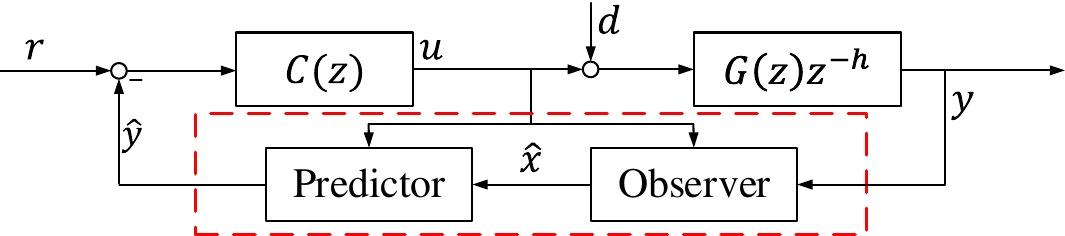}
	\caption{State Predictor with Observer structure.
		\label{fig:state_predictor}}
	\vspace{-10pt}
\end{figure}

Since not all the state $\bm{x}_k$ in the Eq.~\eqref{eq:state_predictor} is measurable, we use an observer to obtain the current angular loop state $\bm{x}_k = [\omega,\ \dot{\omega}]^T$, to construct an observer-predictor structure, as shown in Fig.~\ref{fig:state_predictor}.
Considering the delay-free plant of the Hex-Jet angular loop is a second-order system, an observer~\cite{gao2006active} is designed to estimate $\bm{x}_{k}$:
\begin{equation}
\begin{aligned}
    &\dot{\hat{\bm{x}}} = (\bar{\mathbf{A}} - \bar{\mathbf{L}}\bar{\mathbf{C}})\hat{\bm{x}} + \bar{\mathbf{B}}u + \bar{\mathbf{L}}y \\    
&\bar{\mathbf{A}} = 
\begin{bmatrix}
    0 & 1 \\
    0 & 0
\end{bmatrix}
,
\bar{\mathbf{B}} =
\begin{bmatrix}
    0 \\
    1
\end{bmatrix}
,
\bar{\mathbf{C}} =
\begin{bmatrix}
    1 & 0
\end{bmatrix} \\
&\bar{\mathbf{L}} = 
\begin{bmatrix}
    {\beta_1} & {\beta_2}
\end{bmatrix}
^T
\end{aligned} 
\end{equation}
where $\hat{\bm{x}} = \begin{bmatrix} \dot{\hat{\omega}} & \hat{\omega} \end{bmatrix}^T$ is the variable of the observer, and $\bar{\mathbf{L}}$ is the gain of the observer.
By choosing propper observer gain $\bar{\mathbf{L}}$ such that $\bar{\mathbf{A}} - \bar{\mathbf{L}}\bar{\mathbf{C}}$ is asymptotically stable, then $\hat{\omega} \rightarrow \omega$ and $\dot{\hat{\omega}} \rightarrow \dot{\omega}$.

\section{EXPERIMENTAL VERIFICATION IN QUADROTOR PLATFORM}
\label{sec:quad_experiment_verification}
In this section, we conduct experimental verification to assess and compare the tracking performance and robustness of two predictor-based controllers on a quadrotor platform before implementing the predictor-based controller on our scaled Hex-Jet.
Our controller is implemented in a customized PX4 autopilot flight control software~\cite{meier2015px4}. 
In general, the motor response speed of the quadrotor is sufficiently fast. 
Our system identification results in Fig.~\ref{fig:sys_identification} show that the delay primarily affects the high-frequency response of the control system.
Agile, small aircraft are more sensitive to delays because they require a higher control bandwidth.
Therefore, to emphasize the impact of delay in the control loop, we introduce an additional delay to the output of the rate controller:
\begin{equation}
    u(k) = u(k-h).
\end{equation}
Given that the quadrotor's rate controller operates at a frequency of 400 Hz, the discretized roll rate transfer function used for predictors is
\begin{equation} \label{eq:quad_trans}
    G_{n}(z) = \frac{-0.7549z + 0.9393}{z^2 - 1.8770z + 0.8786},
\end{equation}
and the corresponding state-space equation of Eq.~\eqref{eq:quad_trans} is
\begin{equation} \label{eq:quad_state}
    \mathbf{A}_{d} = \begin{bmatrix}
    ~\ 0.9991 &  0.0023 \\
   -0.6953  &  0.8777
\end{bmatrix},~ \mathbf{B}_{d} = \begin{bmatrix}
    0 \\
    0.0234
\end{bmatrix}.
\end{equation}

Using the identified system model Eq.~\eqref{eq:quad_trans} and Eq.~\eqref{eq:quad_state}, we integrate the predictor-based controller into the roll angular rate controller.

\begin{figure}[h]
	\centering
	\includegraphics[width=1.0\columnwidth]{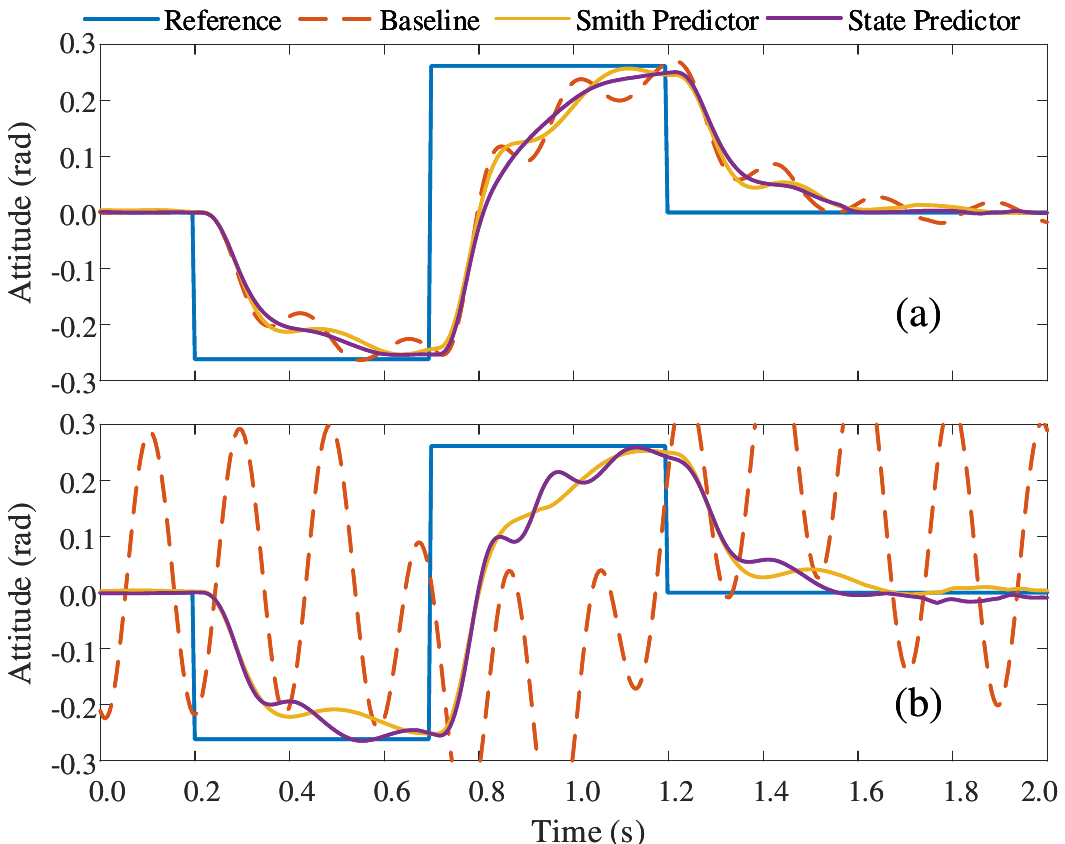}
	\caption{Step Response performance with various additional delay, Fig.~(a) presents the step response with a 5-sample time delay added, Fig.~(b) shows the step response with an additional 10-sample delay.
		\label{fig:quad_step}}
	\vspace{-10pt}
\end{figure}

\begin{figure}[t]
	\centering
	\includegraphics[width=1.0\columnwidth]{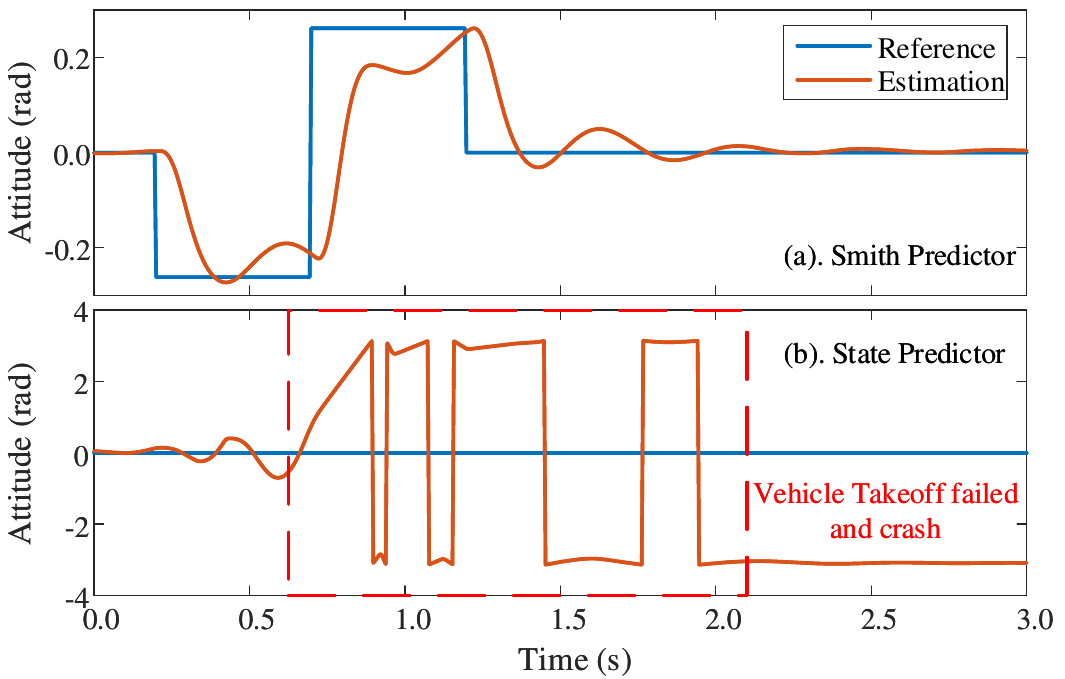}
	\caption{Step Response performance with 25 samples additional delay
		\label{fig:quad_step_25}}
	\vspace{0pt}
\end{figure}

\subsection{Tracking Performance}
\label{subsec:track}
To evaluate the controller's performance to track the attitude reference, a doublet command input is applied to the quadrotor roll attitude controller.
The amplitude of doublet input is 15 degrees, with 0.5 seconds in the positive direction and 0.5 seconds in the negative direction. 
To facilitate the comparison, the doublet input is sent at t = 0.2s, and the duration of data recording is 2s.
The step response results with 5-sample and 10-sample additional delays are illustrated in Fig.~\ref{fig:quad_step}(a) and Fig.~\ref{fig:quad_step}(b), separately.

In the 5-sample additional delay case, we can observe that the baseline controller and two predictor-based controllers exhibit similar dynamic tracking performance. 
All controllers show no distinct overshot in the attitude control loop with the same time constant of the closed-loop system being approximately ${\tau} = 0.1161$ seconds. 
The tracking root mean square error (RMSE) for the three controllers is similarly small, with values of 0.0750, 0.0705, and 0.0717 rad, respectively.
Considering all of the controller gains remain the same, this result indicates that the predictor-based controllers do not improve the response speed. 
Due to the pure delay term in the rate controller~\cite{zhong2006robust}, the baseline controller performs some oscillation when tracking the desired doublet signal. 
The Smith predictor shows less oscillation during tracking and stabilizing, and the state predictor shows no oscillation in the whole experiment.

In the 10-sample additional delay case, we observe that the baseline controller loses stability, with strong oscillation, and fails to track the desired attitude in the experiment, whereas both predictor-based controllers maintain stability with significantly less oscillation.
The two predictor-based controller's RMSE values are 0.0703 and 0.0690 rad, respectively.

The 25-sample additional delay case results are shown in Fig.~\ref{fig:quad_step_25}, to match the actual delay in the Hex-Jet. 
Our quadrotor with the Smith predictor remains stable and completes the step response with minimal overshoot and oscillation.
Compared to cases with 5-sample and 10-sample additional delays, the Smith predictor's RMSE with 25-sample additional delays increases by only 11\%, reaching 0.0787 rad.
However, the state predictor fails to stabilize the vehicle, resulting in strong oscillation during takeoff and eventually causing a crash.

\begin{figure}[t]
	\centering
	\includegraphics[width=1.0\columnwidth]{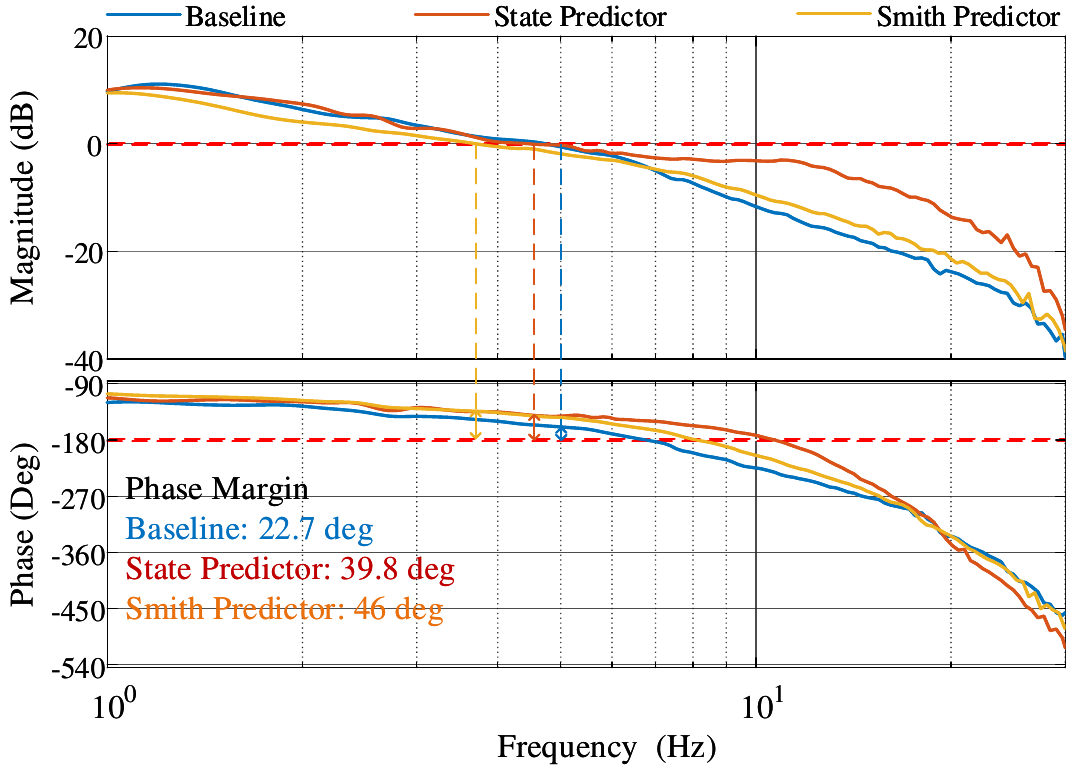}
	\caption{The frequency response results from the controller input $u$ to the rate output $y$ with a 5-sample additional delay. For predictor-based controllers, we use the predicted output $\hat{y}$ instead of $y$.
		\label{fig:quad_frd}}
	\vspace{-10pt}
\end{figure}

\begin{table*}[t]
	\centering
	\caption{Disturbance rejection performance. We use both root mean square error (RMSE) and maximum absolute error (MAE) to evaluate the disturbance rejection performance and the maximum deviation of the rate controller, separately. The red text highlights the differences in RMSE and MAE relative to the baseline controller under the 5-sample delay scenario. Similarly, the blue text denotes the differences in RMSE and MAE compared to the state predictor in the 10-sample delay scenario.}
	\label{tab:quad_dist}
	\resizebox{\textwidth}{!}{
		\begin{tabular}{@{}ccccccc@{}}
			\toprule
			& Baseline & Smith predictor & State predictor & Baseline & Smith predictor & State predictor\\
            & 5-sample-delay & 5-sample-delay & 5-sample-delay & 10-sample-delay & 10-sample-delay & 10-sample-delay\\
            \midrule
			RMSE (rad/s) & 1.4890 & \textbf{0.9512}~\textcolor{red}{(-36.1\%)} & 1.0217~\textcolor{red}{(-31.2\%)} & unstable & \textbf{0.9788}~\textcolor{blue}{(-32.2\%)} & 1.4432\\
                MAE (rad/s) & 6.9428 & \textbf{4.7416}~\textcolor{red}{(-31.70\%)} & 5.9337~\textcolor{red}{(-14.5\%)} & unstable & \textbf{4.7512}~\textcolor{blue}{(-34.6\%)} & 7.2624\\
			\bottomrule
		\end{tabular}
	}
	\vspace{-10pt}
\end{table*}

\begin{figure*}[t]
	\centering
	\includegraphics[width=1.0\textwidth]{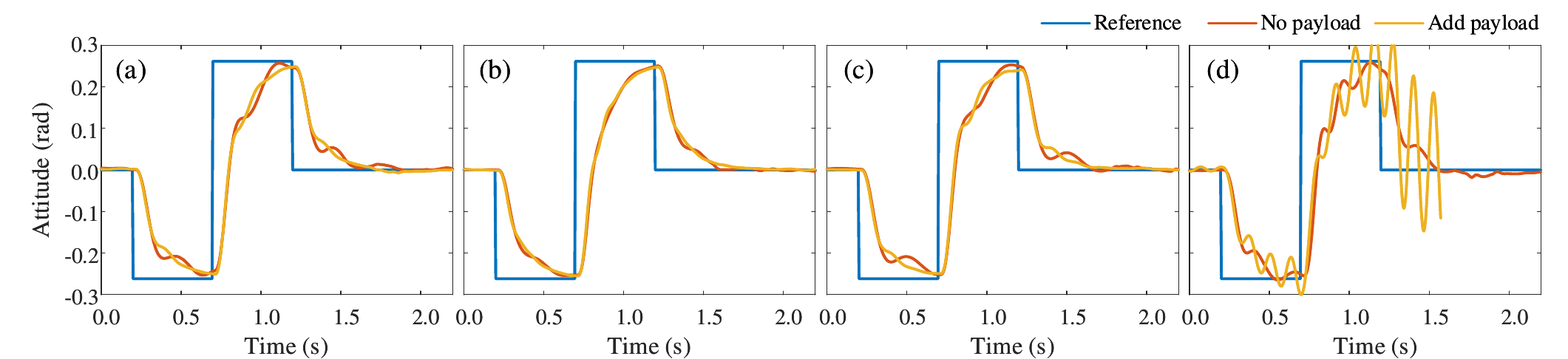}
	\caption{Attitude controller step response performance with/without payload. (a).Smith Predictor with 5 sample delay, (b).State Predictor with 5 sample delay, (c).Smith Predictor with 10 sample delay, (d).State Predictor with 10 sample delay.
	\label{fig:quad_robust_step}}
	\vspace{-10pt}
\end{figure*}

\subsection{Frequency Response}
\label{subsec:frd}
We analyze the enhancement of predictor-based controllers in the frequency domain. 
In the 5-sample additional delay case, we conduct a frequency sweep experiment to obtain the response from controller input $u$ to predicted output $\hat{y}$. 
The transfer function from $u(z)$ to $\hat{y}(z)$ is denoted as the augmented plant $G^{*}(z)$, given by
\begin{equation}
    G^{*}(z) = \frac{\hat{y}(z)}{u(z)}.
\end{equation}
For the baseline controller, we use angular rate output $y$ instead of $\hat{y}$ since there is no prediction for the baseline controller.

The open-loop frequency response results of open-loop transfer function $C(z)G^{*}(z)$ are shown in Fig.~\ref{fig:quad_frd}. 
From the Bode diagrams, we can observe that all the controllers have the same DC gains, approximately 10 dB.
The cutoff frequencies of the baseline controller, the State predictor, and the Smith predictor are 4.74 Hz, 4.49 Hz, and 3.71 Hz, respectively.
The gain margin of the baseline controller, the state predictor, and the Smith predictor are 4.22 dB, 6.13 dB, and 3.18 dB, respectively.
The phase margin of the baseline controller, the State predictor, and the Smith predictor are 22.7 deg, 39.8 deg, and 46 deg, respectively. 
Compared to the baseline controller, the phase margin of the predictor-based controller is increased, thanks to the replacement of the delayed output $y$ with the predicted output $\hat{y}$.
As shown in Fig.~\ref{fig:quad_frd}, it is evident that for frequencies above 5 Hz, the magnitude of the state predictor-based open-loop transfer function deviates significantly from the baseline and the Smith predictor. 
Thus, the state predictor exhibits more model error than the baseline controller and the Smith predictor in the high-frequency region.

\subsection{Disturbance Rejection}
\label{subsec:dist}
In this experiment, we focus on the disturbance rejection performance of the rate controller with different time delays. 
The external doublet step disturbance signal $\bm{M}^{ext}_{b}$, with an amplitude of 0.1 $rad/s^2$, is added to the rate controller output $u$. 
The resulting disturbance rejection performance is summarized in Table.~\ref{tab:quad_dist}. 

Under the 5-sample additional delay condition, the root mean square error (RMSE) of the baseline controller is 1.4890 rad/s, and the maximum absolute error (MAE) is 6.9428 rad/s. In comparison, the Smith predictor has an RMSE of 0.9512 rad/s and an MAE of 4.7416 rad/s, while the state predictor has an RMSE of 1.0217 rad/s and an MAE of 5.9337 rad/s.
Compared to the baseline controller, the Smith predictor and the state predictor have 36.12\% and 31.38\% less RMSE, and 31.7\% and 14.53\% less MAE, respectively.
The Smith predictor and the state predictor exhibit significantly better anti-disturbance performance than the baseline controller. 
The state predictor shows 7.42\% higher RMSE and 25.14\% greater MAE compared to the Smith predictor, suggesting that the Smith predictor outperforms the state predictor in terms of anti-disturbance capabilities.

In the 10-sample additional delay condition, the baseline controller becomes unstable. The RMSE and MAE of the Smith predictor are 0.9788 rad/s and 4.7512 rad/s, respectively, while those of the state predictor are 1.4432 rad/s and 7.2624 rad/s.
The Smith predictor exhibits 32.2\% lower RMSE and 34.6\% lower MAE compared to the state predictor in this case.

In comparison to the case with a 5-sample additional delay, the RMSE and MAE of the Smith predictor with a 10-sample additional delay only increase by 2.9\% and 0.2\%, respectively.
The results show that the disturbance rejection performance of the Smith predictor does not degrade significantly with a larger time delay.
However, the RMSE and MAE of the state predictor increased by 41.3\% and 22.4\%, respectively, compared to the case with a 5-sample additional delay.

In the comparative results of step response and disturbance rejection, we can conclude that the state predictor appears to be more sensitive to modeling errors, external disturbances, measurement errors, and other factors during numerical integration with a significant delay in the time-domain perspective. This observation is consistent with the frequency-domain result showing that the Smith predictor has a larger phase margin than the state predictor under the same time-delay conditions.

\begin{table}[t]
    \tiny
	\vspace{0pt}
	\centering
	\caption{Disturbance rejection performance with/without payload. The red text highlights the differences in RMSE and MAE relative to the baseline.}
	\label{tab:quad_robust}
	\resizebox{\columnwidth}{!}{
		\begin{tabular}{ccc}
			\toprule
			& Smith predictor & Smith predictor \\
                & 5-sample-delay&5-sample-delay w/ Payload \\
                \midrule
                RMSE (rad/s)&  0.9512 & \textbf{0.9726}~\textcolor{red}{(+2.2\%)} \\
                MAE (rad/s) & 4.7416 &\textbf{5.1532}~\textcolor{red}{(+8.7\%)} \\
                \midrule
                & State predictor & State predictor \\
                & 5-sample-delay&5-sample-delay w/ Payload \\
                \midrule
			RMSE (rad/s)&  1.0217 & 1.2679~\textcolor{red}{(+24.1\%)} \\
                MAE (rad/s) & 5.9337 & 7.1853~\textcolor{red}{(+21.1\%)} \\
			\bottomrule
		\end{tabular}
	}
	\vspace{-10pt}
\end{table} 

\subsection{Robustness to Additional Load}\label{subsec:robust}
To more closely match the application environment of Hex-Jet UAV for aerial transportation, we further investigated the impact of modeling errors on the two predictor-based controllers.
The takeoff weight of our quadrotor platform used in the experiment is 594.8g. 
We added an extra load of 230g to the original quadrotor platform, equivalent to 40\% of the vehicle's original takeoff weight. 
We conducted the step-tracking experiment under different additional delays and load conditions.

The result of the Smith predictor and state predictor with 5-sample additional delay is shown in Fig.~\ref{fig:quad_robust_step}(a) and (b), respectively. 
Both predictor-based controllers demonstrate sufficient robustness with the 5-sample additional delay, maintaining similar tracking performance compared to the original model.
The result of the Smith predictor and state predictor with 10-sample additional delay can be found in Fig.~\ref{fig:quad_robust_step}(c) and (d), respectively. 
From Fig.~\ref{fig:quad_robust_step}(d), we observe that with the additional payload, the Smith predictor maintains satisfactory performance. In contrast, the state predictor exhibits serious performance degradation, ultimately leading to control system instability.
These tracking results indicate that modeling error further aggravates the instability of the state predictor. 

Another experiment is conducted to test the disturbance 
rejection of both predictor-based controllers with varying additional delays and payload conditions.
The results are summarized in Table.~\ref{tab:quad_robust}.
The RMSE and MAE of the Smith predictor with payload increased by only 2.2\% and 8.7\%, respectively, whereas the state predictor's RMSE and MAE increased by 24.1\% and 21.1\%, respectively.
The disturbance rejection results indicate that the Smith predictor exhibits greater robustness to modeling errors compared to the state predictor.

Based on the comparative experimental results of two predictor-based controllers in Sec.~\ref{sec:quad_experiment_verification}, it can be concluded that the Smith predictor demonstrates superior disturbance rejection performance and model error robustness when compared to the state predictor. 
In scenarios with larger delays, the state predictor may lose stability, leading to flight failure, whereas the Smith predictor showcases robust performance in cases of substantial delays.

\begin{figure}[t]
	\vspace{0.0cm}
	\centering
	\includegraphics[width=1\columnwidth]{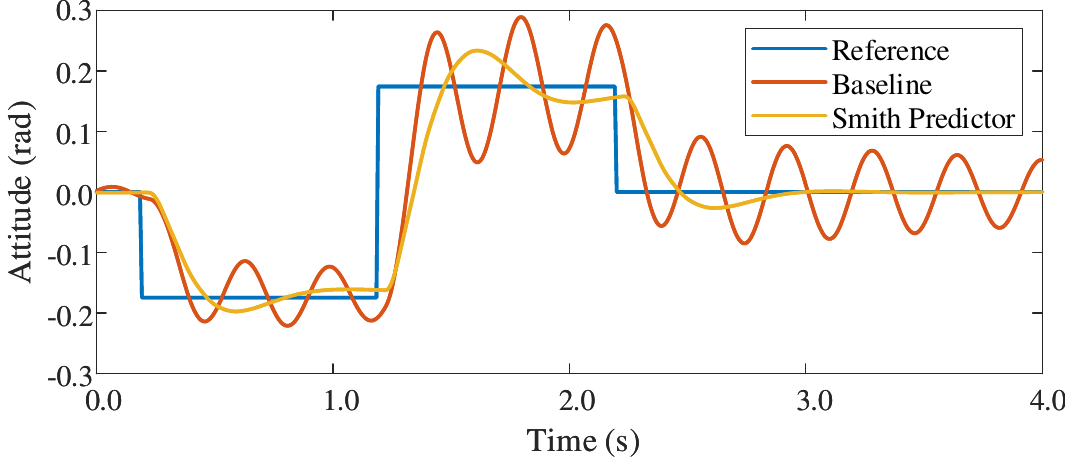}
	\caption{Roll axis attitude step response comparison during the Hex-Jet flight test.
		\label{fig:flight_test_result}}
	\vspace{-10pt}
\end{figure}

\begin{figure}[t]
	\vspace{-0pt}
	\centering
	\includegraphics[width=1\columnwidth]{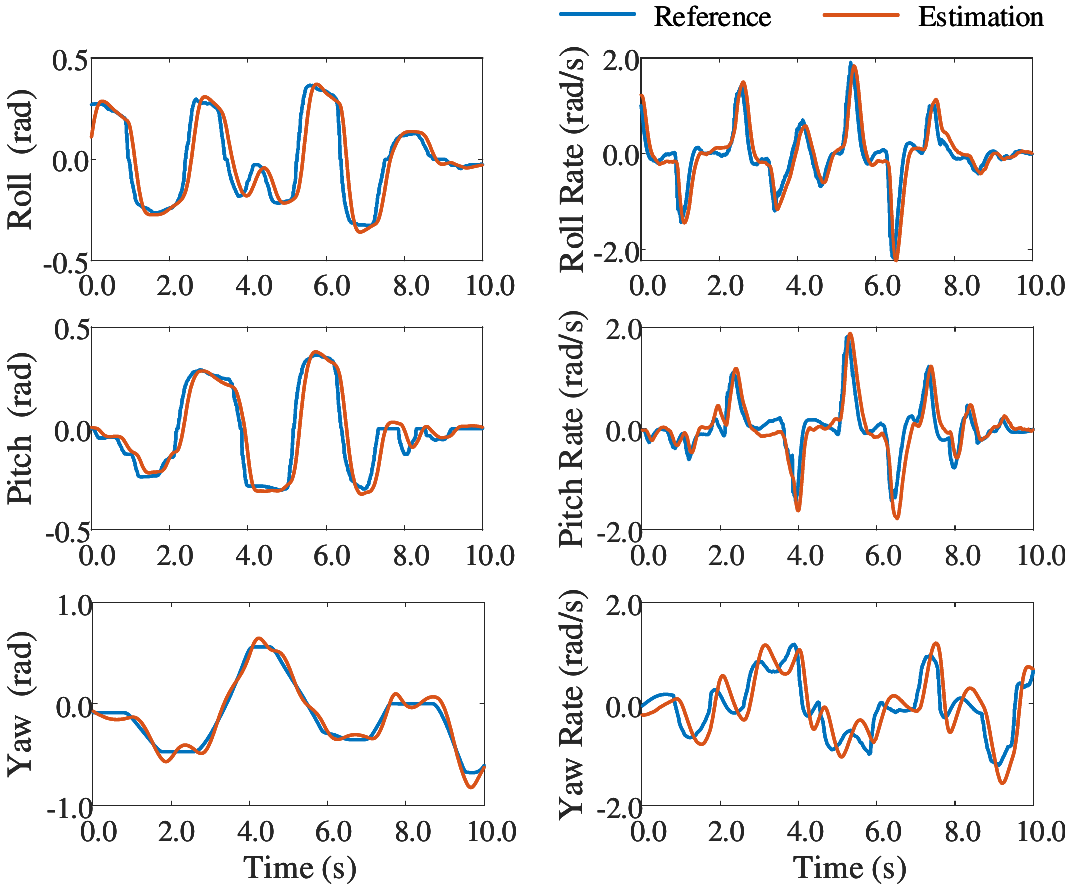}
	\caption{Attitude and rate tracking results of the Hex-Jet in the manual control mode.
		\label{fig:flight_test_pitch_and_yaw_rate_tracking}}
	\vspace{-10pt}
\end{figure}

\section{FLIGHT TEST VERIFICATION ON THE HEX-JET}
\label{sec:flight_test}
Through comparative experiments on a quadrotor UAV, the Smith Predictor-based controller exhibits superior robustness against long-time delays, external disturbances, and model uncertainties. 
Based on these results, we implemented the Smith Predictor-based time delay controller for the flight test verification on the Hex-Jet.
With the refined model~\eqref{eq:quad_trans}, we can verify the effectiveness of our control method on a scaled Hex-Jet.
The sample time of the flight controller is 0.004s. Considering the time delay in our scaled Hex-Jet is sufficiently large (i.e. 25 sample time delay or 0.1s), as well as the challenges of external disturbances, we choose the Smith predictor to validate the flight control performance on the scaled Hex-Jet.

The roll axis attitude step response results are shown in Fig.~\ref{fig:flight_test_result}. 
A doublet attitude setpoint command is given by the flight controller, and the magnitude is 10 deg.
Compared to the baseline controller without a rate prediction process, the Smith predictor-based controller exhibits 42.6\% less overshoot and 64.3\% faster settling time. 
The tracking RMSE and MAE for the Smith predictor are 0.0599 and 0.3810 rad, respectively.
In contrast, the baseline controller failed to stabilize the vehicle during the test with intense oscillation.

We validate the roll, pitch, and yaw axis attitude tracking performance of the scaled Hex-Jet, and the results are shown in Fig.~\ref{fig:flight_test_pitch_and_yaw_rate_tracking}. 
The attitude setpoint signal is given by a remote controller during the test.
During our flight test maneuver, both the roll attitude and rate controller show good tracking performance with large actuator delay, the RMSE of the roll attitude controller and the rate controller is 0.0544 rad and 0.2131 rad/s.
The results also show that the pitch and yaw attitude and rate-tracking performance meet our expectations.
The RMSE and MAE for the pitch and yaw axis attitude control are 0.0470 and 0.049 rad, respectively. The RMSE for the rate control is 0.2103 and 0.3217 rad/s, respectively.

\section{CONCLUSION AND FUTURE WORK}
\label{sec:conclusion}

In this letter, we introduce the Hex-Jet, a turbojet-powered UAV platform that integrates thrust vectoring and differential thrust control. 
Our platform offers a simplified mechanical structure compared to the uniaxial and biaxial thrust vectoring mechanisms used in conventional jet-powered UAVs. 

To address the turbojet engine's time delay challenge in roll axis control, we design two predictor-based controllers based on the identified frequency model. 
Through comparative experiments on a quadrotor platform, we showcase the Smith predictor's superior performance over the state predictor, particularly in disturbance suppression and model uncertainty handling with a large time delay.
Subsequently, we validate the efficacy of the Smith predictor-based rate controller on a scaled Hex-Jet through flight tests. 
With the uniaxial thrust vectoring and the Smith predictor-based roll rate controller, our prototype can realize robust full attitude control with satisfactory performance results.

In future work, we plan to verify the performance of this improved controller on a full-scale Hex-Jet powered by turbojet engines.

\bibliography{Reference} 
\end{document}